\documentclass[lettersize,journal]{IEEEtran}
\usepackage{amsmath,amsfonts}
\usepackage{algorithm}
\usepackage[noend]{algpseudocode}
\usepackage{array}
\usepackage[caption=false,font=normalsize,labelfont=sf,textfont=sf]{subfig}
\usepackage{textcomp}
\usepackage{stfloats}
\usepackage{url}
\usepackage{verbatim}
\usepackage{graphicx}
\usepackage{cite}
\usepackage{multirow}
\usepackage{tabularx}
\usepackage{colortbl}
\usepackage[hang]{footmisc}
\usepackage[dvipsnames]{xcolor}
\usepackage{xcolor}      
\usepackage{booktabs}   
\usepackage{amsmath}  
\usepackage{amssymb}

\definecolor{c1}{HTML}{FF8C00}
\definecolor{c2}{HTML}{E57373}
\definecolor{c3}{HTML}{FFB2B2}
\definecolor{c4}{HTML}{FFE3E3}
\definecolor{c5}{HTML}{FFB400}
\makeatletter
\renewcommand{\ALG@beginalgorithmic}{\footnotesize}

\makeatother

\begin{document}
\raggedbottom

% \begin{document}

\title{TEn-CATG:~Text-Enriched Audio-Visual Video~Parsing with Multi-Scale Category-Aware Temporal~Graph}

\author{Yaru Chen, Faegheh Sardari, Peiliang Zhang, Ruohao Guo, Yang Xiang,\\
Zhenbo Li,~\IEEEmembership{Member,~IEEE}, and Wenwu Wang,~\IEEEmembership{Senior Member,~IEEE}%
\thanks{Y. Chen, F. Sardari, Y. Xiang and
W. Wang are with the Centre for Vision, Speech and Signal Processing
(CVSSP), University of Surrey, Guildford, UK. 
Email: \{yaru.chen, f.sardari, yang.xiang, w.wang\}@surrey.ac.uk. 
P. Zhang is with the School of Computer Science and Artificial Intelligence, 
Wuhan University of Technology, Wuhan, China. 
Email: cheungbl@ieee.org. 
R. Guo is with the National Key Laboratory of General Artificial Intelligence, 
School of Intelligence Science and Technology, Peking University, Beijing, China. 
Email: ruohguo@stu.pku.edu.cn. 
Z. Li is with the College of Information and Electrical Engineering, 
China Agricultural University, Beijing, China.}
}

\maketitle
\begin{abstract}
Audio-visual video parsing (AVVP) aims to detect event categories and their temporal boundaries in videos, typically under weak supervision. Existing methods mainly focus on (i) improving temporal modeling using attention-based architectures or (ii) generating richer pseudo-labels to address the absence of frame-level annotations. However, attention-based models often overfit noisy pseudo-labels, leading to cumulative training errors, while pseudo-label generation approaches distribute attention uniformly across frames, weakening temporal localization accuracy.
To address these challenges, we propose TEn-CATG, a text-enriched AVVP framework that combines semantic calibration with category-aware temporal reasoning. More specifically, we design a bi-directional text fusion (BiT) module by leveraging audio-visual features as semantic anchors to refine text embeddings, which departs from conventional text-to-feature alignment, thereby mitigating noise and enhancing cross-modal consistency. Furthermore, we introduce the category-aware temporal graph (CATG) module to model temporal relationships by selecting multi-scale temporal neighbors and learning category-specific temporal decay factors, enabling effective event-dependent temporal reasoning. 
Extensive experiments demonstrate that TEn-CATG achieves state-of-the-art results across multiple evaluation metrics on benchmark datasets LLP and UnAV-100, highlighting its robustness and superior ability to capture complex temporal and semantic dependencies in weakly supervised AVVP tasks.
\end{abstract}


\section{Introduction}
\begin{figure}[t] 
\centering
\includegraphics[width=0.43\textwidth]{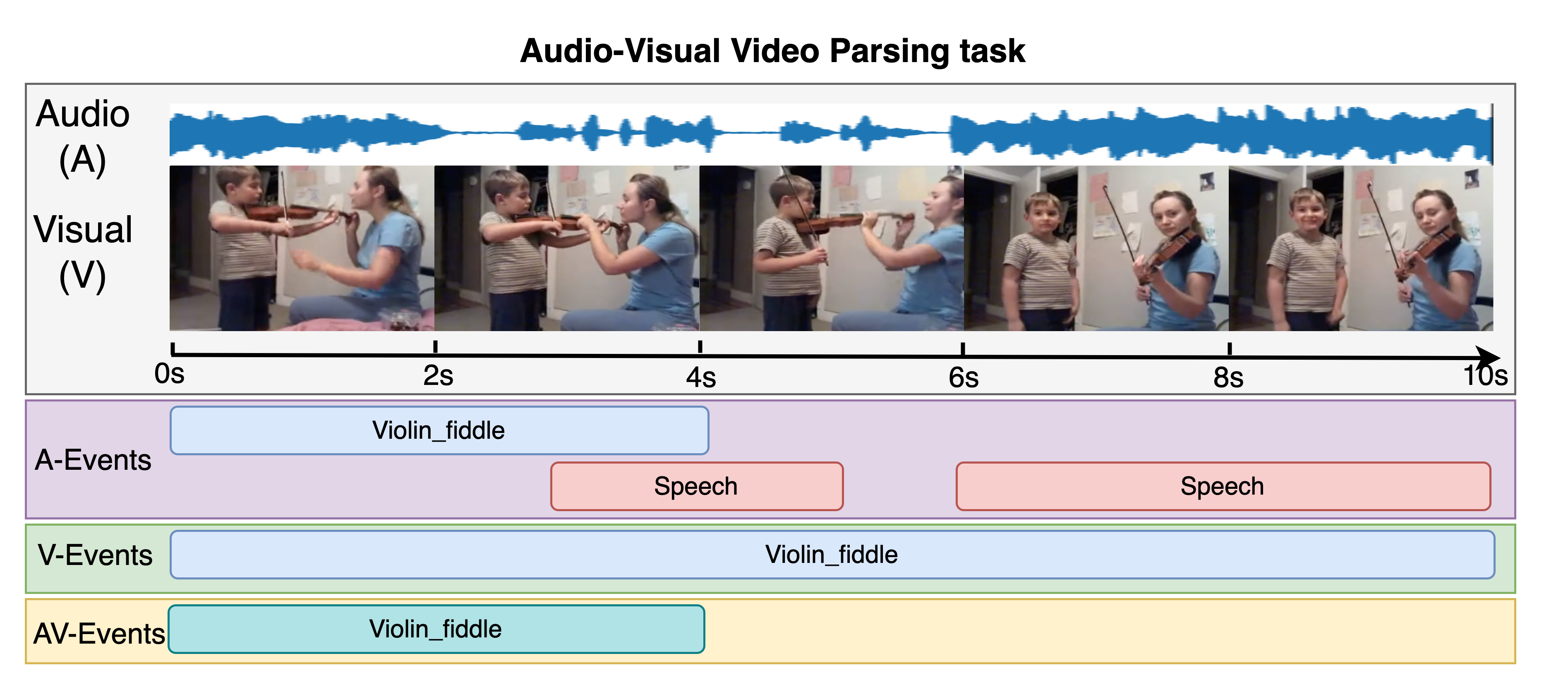} % 调整图片宽度
    \caption{Illustration of the AVVP task. An event may occur in one modality, or across different modalities, and at different times.} 
    \label{fig:motivation}
\end{figure}

\IEEEPARstart{T}{he} audio visual video parsing (AVVP) task aims to not only identify what and when events occur, but also determine which modality~-audio, visual, or both- is responsible for detecting the event~\cite{tian2020unified}. As shown in Fig.~\ref{fig:motivation}, a key characteristic distinguishes AVVP from other audio-visual tasks~\cite{xue2021audio,liu2024bavs,guo2025audio,che2021constrained} is the inherent temporal asynchrony between modalities: an event may exist in one modality but not in another, or the event may occur at different times across modalities. 
This modality-specific and temporally misaligned characteristics make fine-grained temporal modeling essential for accurately localizing the time of events occurring and assigning modality tags. In addition, AVVP is often performed under a weakly supervised setting, where only video-level labels are available during training, especially when frame-level labels are infeasible or expensive to obtain. This further increases the challenge of learning temporal information and cross-modal interactions without segment-level supervision. Hence, how to fully explore the semantic relationships between video frames with limited data is a challenge for the AVVP task. 

Several approaches have been proposed to address these challenges in two aspects. One focuses on using attention-based encoders that model temporal dependencies across segments~\cite{tian2020unified,wu2021exploring,jiang2022dhhn,yu2022mm}. The other explores the generation of pseudo-labels, often with the help of pre-trained audio-text or vision-language models, for the frame missing annotations, and guides cross-modal learning~\cite{lai2023modality,zhou2024label,jiang2024resisting,wang2025link,zhao2025text}. Although both directions have shown effectiveness in addressing different aspects of the AVVP task, they still have limitations. Firstly, these approaches treat segments in isolation, either relying on rigid temporal structures or assuming pseudo-labels to be semantically accurate. In practice, pseudo-labels are often noisy or inaccurate, and may propagate errors when used indiscriminately. Moreover, temporal modeling without semantic awareness can be limited for capturing meaningful inter-segment relationships crucial for consistent event reasoning. 

To address both issues in a unified way, we propose TEn-CATG, a text-enriched (TEn) AVVP framework that integrates a bi-directional text (BiT) fusion module and a category-aware temporal graph (CATG) module. The BiT module is designed to address the problem of pseudo-label noise. More specifically, it first derives text embeddings from pseudo labels via a contrastive language audio pretraining (CLAP)~\cite{wu2023large} or contrastive language image pretraining (CLIP)~\cite{radford2021learning} model, then performs co-attention with audio-visual features and injects global modality semantics to calibrate the embeddings, yielding segment-level representations with reliable cross-modal alignment. The CATG module addresses the limitation of temporal modelling in previous designs, by constructing a multi-scale temporal graph, where each segment adaptively selects temporal neighbors according to its predicted event category, capturing event-specific durations and patterns. In short, BiT ensures cross-modal semantic reliability, while CATG provides event-aware temporal reasoning. They together form a coherent framework that explicitly addresses the limitations in existing AVVP methods, namely, noisy supervision and inflexible temporal modeling.

This work is an extended version of the text-enhanced multi-hop temporal graph (TeMTG) model~\cite{chen2025temtg}, presented in our previous conference paper. TeMTG used a fixed-hop design and lacked category-specific modeling, and only a simple multi-layer perceptron~(MLP)~\cite{taud2017multilayer} to fuse audio-visual features with text embeddings. In our new TEn-CATG model, however, the rigid graph is replaced by the CATG module, which learns hop preferences and decay factors from the predicted event distribution, giving each segment an event-specific temporal receptive field. We also design the BiT module to calibrate pseudo-text embeddings for effectively suppressing label noise. In addition, we conducted comprehensive ablation and visualization experiments on additional datasets to demonstrate the effectiveness and robustness of our new model.

The main contributions of this paper can be summarized as follows.

\begin{enumerate}
    \item To our knowledge, this is the first work to introduce a temporal graph neural network into the AVVP task, where we propose the CATG module that captures segment-specific multi-scale dependencies by adaptively selecting temporal hops based on event semantics.
    \item We propose the BiT module, which enables the pseudo labels to dynamically adapt to modality-specific semantic contexts, thereby mitigating semantic noise. This shifts the design paradigm from conventional fixed text-to-feature alignment to feature-aware text calibration.
    \item We perform comprehensive experiments on two benchmark datasets, LLP and UnAV-100, respectively. The results show that our framework achieves state-of-the-art~(SOTA) performance in multiple evaluation metrics.
\end{enumerate}

\begin{figure*}[h]
    \centering    \includegraphics[width=1.0\textwidth]{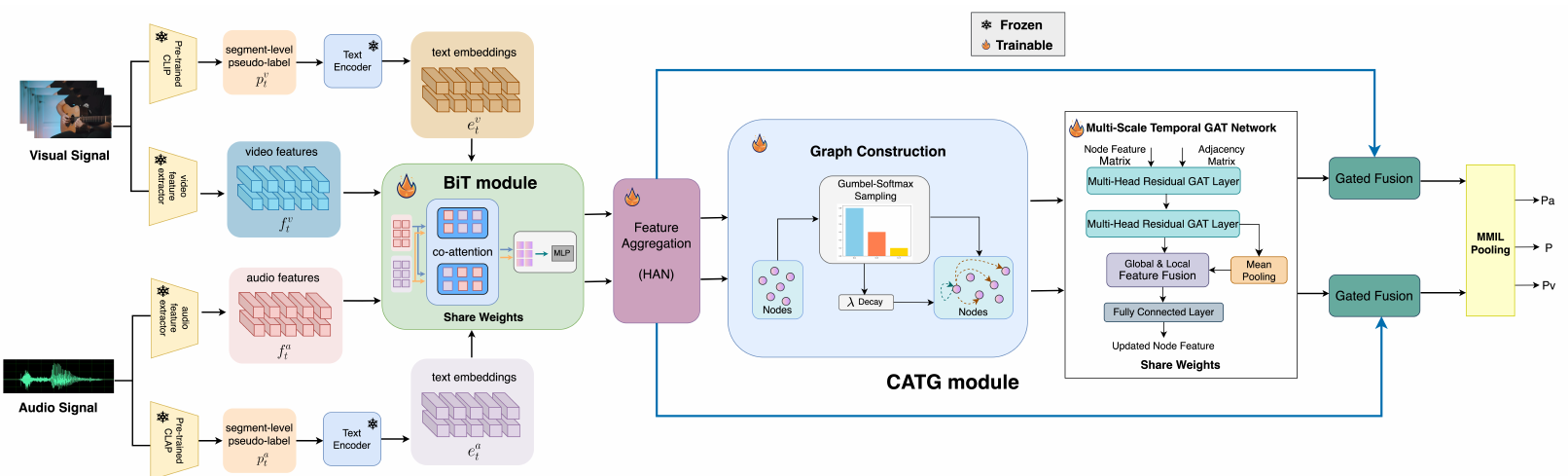} % 调整图片宽度
    \caption{Overview of our model. Segment-level audio/visual features are extracted by frozen CLAP/CLIP, and segment-level text embeddings are generated from pseudo labels. The BiT module is introduced for feature-aware text calibration via co-attention, then the features are aggregated with HAN and enhanced with CATG, a multi-scale temporal graph (i.e. residual GAT). Finally, a gated fusion and MMIL pooling produce audio (Pa), visual (Pv), and joint (P) predictions.}
    \label{fig:model}
\end{figure*}
\vspace{-4mm}
\section{Related Work}
\subsection{Audio-Visual Video Parsing}
The AVVP task aims to jointly detect the occurrence, timing, and type of events using weak video-level annotations. The task was introduced by Tian et al.~\cite{tian2020unified}, along with the look, listen and parse~(LLP) dataset and a baseline framework that integrates a hybrid attention mechanism~(HAN) with a multimodal multiple instance learning~(MMIL) based pooling strategy to extract relevant segment-level cues. Subsequent research efforts focus mainly on two complementary directions. The first is on advancing feature representation backbones, such as the various strategies to enhance intra- and inter-modal semantic consistency~\cite{wu2021exploring,gao2023collecting,sardari2024coleaf}, and the hierarchical or multi-scale temporal structures to capture event durations across time~\cite{yu2022mm,zhang2023multi}. The second one is to improve supervision signal by generating pseudo-labels, such as using the pre-trained audio-text or vision-language models (e.g., CLAP~\cite{wu2023large} and CLIP~\cite{radford2021learning}) to infer frame-level labels from video-level tags~\cite{zhou2024advancing,zhou2024label,jiang2024resisting}, and incorporating text embeddings generated by the pre-trained large language models~\cite{wang2025link,zhao2025text}. %These efforts have advanced the AVVP task in various directions. 
However, these approaches rely on static temporal structures or treat pseudo-labels as reliable ground-truth without considering their semantic noise. In contrast, we design the BiT module to calibrate noisy pseudo labels through modality-aware conditioning, and the CATS module to learn the category-dependent multi-scale graphs, thereby offering reliable supervision and flexible temporal modeling under weak labels.
\vspace{-4mm}
\subsection{Text-guided Audio-Visual Learning}
Recent studies increasingly leverage text as a high-level semantic modality to guide audio-visual learning. Text provides explicit supervision, enabling cross-modal alignment and enhancing representation learning~\cite{liu2023sounding}. As examples, texts have been integrated into pretraining frameworks via masked modeling, or tri-modal learning to align audio, visual, and textual modalities~\cite{fan2024text,zhu2023vatlm}. This improves generalization across modalities and supports semantic consistency~\cite{tan2023language,mahmud2024t}. Other methods adopt large language models or pre-trained text embeddings to ground audio-visual content in semantic space~\cite{chowdhury2024meerkat,guzhov2022audioclip}. These approaches enable spatial-temporal localization and content-aware learning without dense labels. In addition, textual queries have been used for source localization and event understanding, aligning natural language descriptions with audio-visual signals via attention-based fusion~\cite{koepke2022audio,zhao2024text}. These methods rely on static text embeddings or unidirectional alignment (text to feature), which limits adaptability to ambiguous or evolving events. Moreover, reliance on annotated queries introduces bias and noise in weakly-supervised settings. In contrast, our BiT module produces context-adaptive, noise-robust alignments without relying on static text tokens or additional annotation queries.
\vspace{-4mm}
\subsection{Graph-based Audio-Visual Learning}
Graph-based models have emerged as a powerful paradigm for audio-visual learning, due to their flexibility in capturing cross-modal relationships, temporal dependencies, and structural semantics~\cite{ektefaie2023multimodal,zhang2025subgraph}. Previous works can be divided into three main categories. The first is to model temporal information using the graph structure~\cite{wang2024multimodal,nie2020c,min2022learning,montesinos2022vovit}, employing hand-crafted temporal edge connections for short-term temporal modeling~\cite{wang2024multimodal,nie2020c}, or using spatial-temporal graph neural networks to capture long-range dependencies~\cite{min2022learning,montesinos2022vovit}. The second is to leverage the cross-modal interaction by constructing multimodal graphs with herterogeneous nodes and edges, thus aligning audio and visual semantics through attention or dynamic edge construction~\cite{liu2023tmac,nguyen2024hig}. The third is to customize a graph structure for a variety of audio-visual tasks such as emotion recognition~\cite{jia2021hetemotionnet}, question answering~\cite{zhao2024heterogeneous}, and event localization~\cite{liu2025audio}. These approaches often incorporate semantic priors, scene attributes, or label co-occurrence patterns into the graph structure, enabling task-specific reasoning. Unlike hand-crafted or task-specific graph designs, our CATG model selects adaptively the jump size based on the predicted event class.

\section{Proposed Method}
\subsection{Problem Formulation}
In the weakly supervised AVVP task, each input video $\mathcal{S}$ is divided into $T$ segments, denoted as \(\mathcal{S}=\{\mathbf{a}_{t}, \mathbf{v}_{t}\}_{t=1}^T\), where $\mathbf{a}_t\in\mathbb{R}^d$ and $\mathbf{v}_t\in\mathbb{R}^d$ represent the audio and visual features of the $t$-th segment, with $d$ being the corresponding feature dimensions. The task has two main objectives, one is to identify three types of events: audio-only events $\mathbf{y}_t^a \in \lbrace 0,1 \rbrace^C$, visual-only events $\mathbf{y}_t^v \in \lbrace 0,1 \rbrace^C$, and audio-visual events $\mathbf{y}_t^{av} \in \lbrace 0,1 \rbrace^C$, where $C$ is the total number of event categories. Each label vector is multi-hot, allowing multiple events to co-occur at the same segment. In addition, audio-visual event is considered to occur only when the audio event and visual event happen simultaneously, which means \(\mathbf{y}_t^{av} = \mathbf{y}_t^a \odot \mathbf{y}_t^v\), where $\odot$ represents the logical and operation. The second objective is to localize the temporal boundaries of each event. During training, only weak (i.e. video-level) labels $\mathbf{y} \in \lbrace 0,1 \rbrace^C$ are available.
\vspace{-4mm}
\subsection{Overview of The Proposed Framework}
We build upon CoLeaF~\cite{sardari2024coleaf} as our baseline and introduce a novel multimodal optimization framework that integrates bi-directional text fusion and category-aware temporal structure modeling. Our framework establishes a mutually reinforcing pipeline where semantics-aware feature calibration directly informs adaptive temporal reasoning, enabling a more coherent and context-driven parsing of audio-visual events.
\
As shown in Fig.~\ref{fig:model}, we first extract audio and visual features using pre-trained CLAP~\cite{wu2023large} and CLIP~\cite{radford2021learning} encoders, and generate segment-level text embeddings from pseudo labels provided by VALOR~\cite{lai2023modality}. Then, the bi-directional text fusion (BiT) module~(Section~\ref{BiT}) is designed to improve the semantic alignment between the modality features and the text features. The fused features are then processed by a HAN~\cite{tian2020unified} with self- and cross-attention. To model event-specific temporal dependencies, we also introduce the CATG module~(Section~\ref{CATS}) that builds a learnable graph across segments. A gated fusion module adaptively integrates the outputs from HAN and CATG. Finally, MMIL pooling~\cite{tian2020unified} is applied to produce audio, visual, and joint predictions. Similar to CoLeaF~\cite{sardari2024coleaf}, we insert our BiT and CATG modules into both the anchor and reference branches for feature aggregation.
\vspace{-4mm}
\subsection{Bi-Directional Text Fusion}\label{BiT}
To improve the semantic alignment between the features of audio-visual modalities and the supervision information from weak event labels, we introduce the BiT module that enables fine-grained interaction between audio and visual features and textual guidance at segment level. 
Before applying the BiT module, we first utilize the segment-level pseudo labels provided by VALOR~\cite{lai2023modality}, where each label is a 25-dimensional binary vector indicating the presence~($1$) or absence~($0$) of event categories. For each segment, we identify the active event classes~(i.e., entries with value $1$) and generate the corresponding textual prompts to construct semantic embeddings for both audio and visual modalities. For that, we design different prompt templates. For audio modality, we use the sentence ``This is the sound of [event]''; for visual modality, we use the sentence ``This is a photo of [event]''. If no event is active for a given segment~(i.e., all-zero vector), the prompt ``There is no event in this segment'' is used for both modalities. 

\begin{figure}[t] 
\centering
\includegraphics[width=0.33\textwidth]{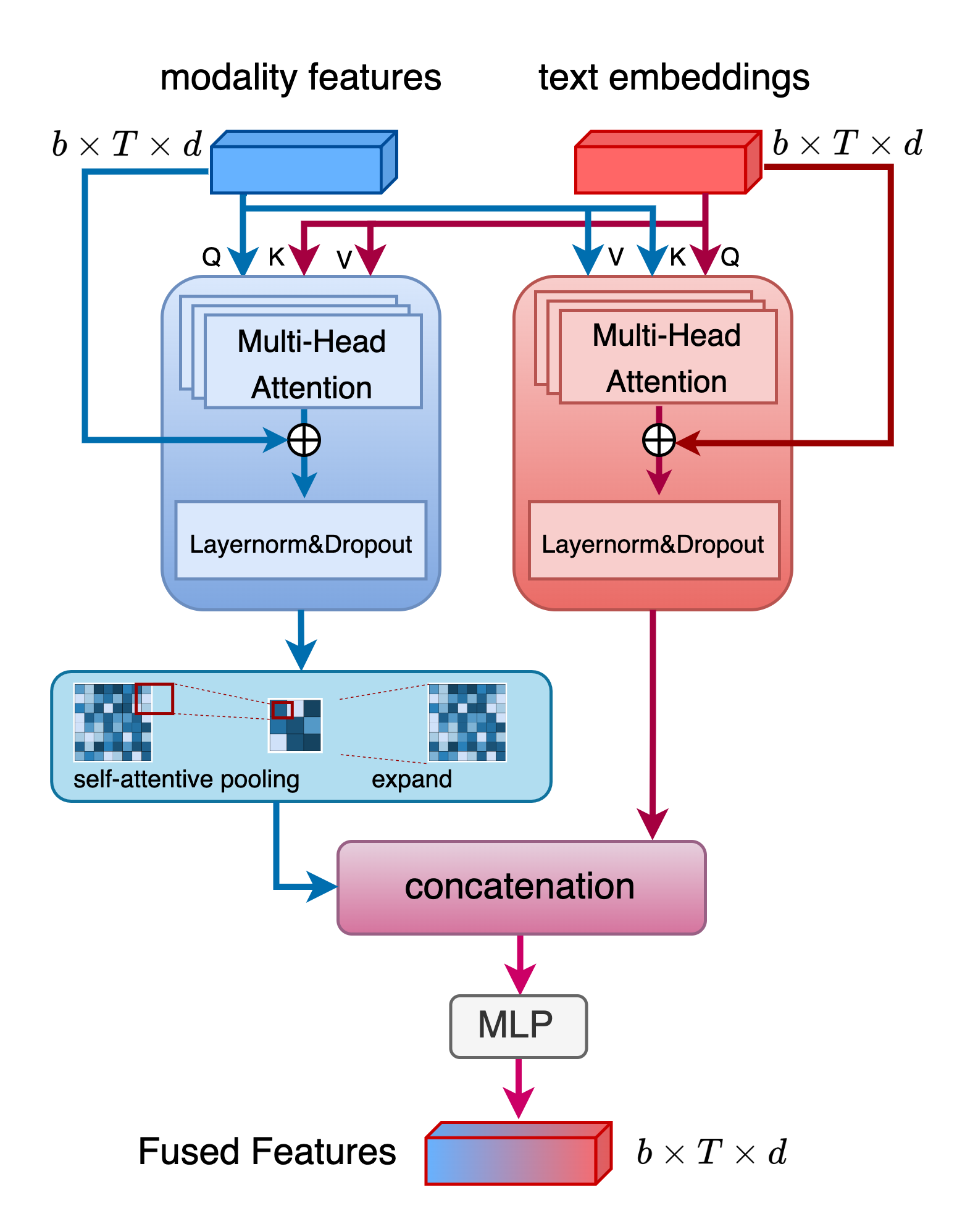} % 调整图片宽度
    \caption{Overview of the BiT module. The modality features and text embeddings interact via bi-directional attention, followed by semantic pooling and fusion through an MLP.} 
    \label{fig:BiT}
\end{figure}

We convert the segment-level pseudo labels into text embeddings using the frozen text encoders from the pre-trained CLAP~\cite{wu2023large} and CLIP~\cite{radford2021learning}. Then, we use BiT to fuse the modality features~$\mathbf{M}\in\mathbb{R}^{b\times T\times d}$ and text embeddings~$\mathbf{E}\in\mathbb{R}^{b\times T\times d}$, where $b$ represents batch size. As shown in Fig.~\ref{fig:BiT}, the bidirectional cross-attention mechanisms are first applied to fuse the modality features and the text embeddings. Each cross-attention output is processed by performing a dropout operation, followed by residual addition and layer normalization to stabilize training. These updates are as follows:
\begin{equation}
\label{MHAf2t}
\mathbf{M}^{'} = \text{LN}(\mathbf{M}+\text{Dropout}(\text{MHA}(\mathbf{M},\mathbf{E},\mathbf{E})))
\end{equation}
\begin{equation}
\label{MHAt2f}
\mathbf{M}^{'} = \text{LN}(\mathbf{M}+\text{Dropout}(\text{MHA}(\mathbf{E},\mathbf{M},\mathbf{M})))
\end{equation}
where \text{LN} represents layernorm operation, and \text{MHA} is multi-head attention.

Previous works, e.g. ~\cite{zhou2024advancing,zhou2024label,jiang2024resisting}, often rely only on the supervision of pseudo-labels for feature-text alignment. However, they are prone to noise and semantic drift. We argue that the modality features contain more reliable semantic cues. As ``semantic anchors'', modality features can be better leveraged to refine text representations and mitigate the noise inherent in pseudo-labels. Hence, we utilize a self-attentive pooling mechanism to extract global semantic information from modality features. Self-attentive pooling is performed as follows. Firstly, we calculate an attention score $s$ for each segment through a linear projection. Concretely, the attention weights $\boldsymbol{\alpha}_t$ for each segment are computed as:
% \begin{equation}
% \label{attention score}
% s = W_{att} F^{'}_m \in\mathbb{R}^{b\times T\times 1}
% \end{equation}
\begin{equation}
\label{attention weights}
\boldsymbol{\alpha} _t = \frac{\text{exp}({\mathbf{w}^{\top}_{att}}\mathbf{M}^{'}_{t}) }{ {\textstyle \sum_{i=1}^{T}} ({\mathbf{w}^{\top}_{att}}\mathbf{M}^{'}_{i})  } \in \mathbb{R}^{B}
\end{equation}
where $\mathbf{M}^{'}_t$ means the features corresponding to each segment $t$ in $\mathbf{M}^{'}$, and $\mathbf{w}_{att}\in\mathbb{R}^{d}$ is a learnable projection parameter. Afterwards, the global semantic vector $\mathbf{g}_m$ is computed as the weighted sum of the segment-level features:
\begin{equation}
\label{global vector}
\mathbf{G} = \sum_{t=1}^{T}\alpha _t\cdot \mathbf{M}^{'}_t \in \mathbb{R} ^{B\times }
\end{equation}
 This global representation $\mathbf{G}$ can capture the salient modality-specific semantics, which are subsequently expanded and fused with text embeddings. Finally, the concatenated features are fed into an MLP~\cite{taud2017multilayer} to project the fused representations back to the original feature with dimension~$d$. This enables effective integration of modality-specific semantics and refined textual guidance, producing enhanced segment-level features for downstream event parsing. Note that the BiT module is incorporated into the audio and visual branches separately, while their parameters are shared, which can improve cross-modal training.
\vspace{-4mm}
\subsection{Category-Aware Multi-Scale Temporal Graph}\label{CATS}
To model event-specific temporal structures, we propose the CATG module, which dynamically constructs the temporal graph guided by predicted category information.

\subsubsection{Category-Guided Temporal Graph Construction}
\label{CATGC}
We aim to construct a category-aware multi-scale temporal graph $\mathcal{G} = (\mathcal{N}, \mathcal{E})$ that dynamically models event-specific temporal structures through a learnable category-guided connection, where $\mathcal{N}$ is the set of nodes and $\mathcal{E}$ is the set of temporal edges. In this graph, we set each segment as a node, and the temporal edges between the nodes are dynamically determined based on the predicted event categories. Let $\mathbf{P}\in \mathbb{R}^{B\times T\times C}$ denote the predicted category probabilities for each segment. For each node, we project $\mathbf{P}$ into a set of preference scores $\mathbf{H}\in\mathbb{R}^{B\times T\times K}$ over a candidate hop set $\mathcal{K} = \left \{  1,2,...,K\right \} $, via:
\begin{equation}
\label{for picking k}
\mathbf{H}=\mathbf{P}\mathbf{W}_K\in\mathbb{R}^{b\times T\times K}
\end{equation}where $\mathbf{W}_K\in\mathbb{R}^{C\times K}$ is a learnable matrix that maps category probabilities to temporal hop preferences.
Then we apply the Gumbel-Softmax~\cite{jang2016categorical} sampling to compute the soft selection scores for candidate hops.
Gumbel-Softmax is a differentiable approximation to categorical sampling, enabling discrete hop selection while allowing end-to-end gradient-based optimization. Using that, the model is enabled to dynamically and flexibly select temporal neighbors during training without breaking the differentiability, thus capturing diverse temporal structures more effectively. 
We compute the hop sampling distribution $s_t$ for each:
\begin{equation}
\label{Gumbel softmax}
\mathbf{s}_{t,i} = \frac{exp((\mathbf{h}_{t,i}+\mathbf{n}_{t,i}))/\tau }{ {\textstyle \sum_{j=1}^{K}}exp((\mathbf{h}_{t,j}+\mathbf{n}_{t,j}))/\tau }\in\mathbb{R}^B
\end{equation}
where $\mathbf{n}_{t,i}$ is the Gumbel noise and $\tau$ is the temperature parameter controlling the sampling sharpness. 
Based on these scores, we denote the selected hop indices as $\mathcal{L}=\left\{l_1,l_2,...,l_k\right\}$, Here, $k$ is a hyper-parameter that specifies the number of hops selected from the candidate hop set 
$\mathcal{K}$.

To further model the temporal sensitivity across different categories, we introduce a learnable decay $\mathbf{Q}$ computed as:
\begin{equation}
\label{Gumbel softmax}
\mathbf{Q} = \mathbf{Pw}_\lambda\in\mathbb{R}^{B\times T}
\end{equation}
where $\mathbf{w}_\lambda\in \mathbb{R}^C$ is a learnable decay vector. The design of $\mathbf{Q}$ follows the intuition that, in most audio-visual events, the closer the segments are in time, the more likely they are semantically correlated, although such a correlation tends to decay as the temporal distance increases. In addition, different events exhibit different temporal sensitivities. For example, ``dog barking" is often a transient event, with strong local correlations, while ``lawn mower" is a persistent event, with long temporal dependencies. By learning different decay factors, the model can better adapt to the temporal dynamics of different events. 

After selecting the temporal neighbors, we compute the edge weight for each connection. Given a selected hop distance $l_i\in \mathcal{L}$ with a soft selection score $\mathbf{s}$, we compute the final edge weight using an exponential decay term:
\begin{equation}
\label{edge weight}
\mathbf{w}_{t,i} = \mathbf{s}_{t,i} \odot \text{exp}(-{\mathbf{q}}_t l_i)
\end{equation}
where $\odot$ denotes element-wise multiplication and $\mathbf{q}_t\in \mathbb{R}^{B}$ is the $t$-th column vector of $\mathbf{Q}$. This allows the model to assign lower weights to longer-range temporal connections, while retaining differentiability. 

In summary, each node not only has self-loop to reserve information of the node, but selects future neighbors based on the sampled hops and decay-modulated weights. The whole process of graph construction is summarized in Algorithm~\ref{alg:cats-construct}.

\subsubsection{Multi-Scale Graph Propagation}
Following the graph construction described in Section~\ref{CATGC}, we adopt a multi-scale temporal aggregation network to effectively propagate and refine temporal features over the constructed graph. This component aims to capture hierarchical contextual information while preserving local semantics for each node. As shown in Fig.~\ref{fig:model}, firstly, we input the node features $\mathbf{X}\in \mathbb{R}^{B \times T\times d}$, and the weighted adjacency matrix $\mathbf{A}$ with edge weight matrix $\mathbf{W}$ with elements $\mathbf{w}_{t,i}$ calculated in \ref{edge weight}. Then, we perform two layers of residual graph attention to refine the node representations: 
\begin{equation}
\label{gat}
\mathbf{Z}^{(1)}=\text{GAT}_1(\mathbf{X},\mathbf{A},\mathbf{W})+R_1(\mathbf{X})\in\mathbb{R}^{b\times T\times d}
\end{equation}
\begin{equation}
\label{gat}
\mathbf{Z}^{(2)}=\text{GAT}_2(\mathbf{Z}^{(1)},\mathbf{A},\mathbf{W})+R_2(\mathbf{Z}^{(1)}) \in \mathbb{R}^{b\times T\times d}
\end{equation}
where $\text{GAT}_1$ and $\text{GAT}_2$ are TransformerConv~\cite{velivckovic2018graph,fey2019fast} with four attention heads, in which the edge weights $\mathbf{W}\in \mathbb{R}^{b\times T\times T}$ are passed as one dimensional edge attributes. Specifically, $\mathbf{Z}^{(1)}$ and $\mathbf{Z}^{(2)}$ are the updated node features after each TransformerConv layer, and $R_1$ and $R_2$ denote the residual mappings to align the feature dimensions if necessary.

Afterwards, we apply the mean pooling over all nodes in the same video to incorporate global context, the global vector $g\in\mathbb{R}^{b\times d}$ is then broadcasted and fused with local features by using a gated fusion mechanism:
\begin{equation}
\label{gated fusion}
% x_g = \sigma (W_g[h^{(2)}\left |  \right | g])\cdot h^{(2)}+(1-\sigma (W_g[h^{(2)}]\left |  \right | g))\cdot g
\mathbf{\Tilde{X}} 
= \sigma\!\left(\mathbf{\Tilde{W}} \,[\mathbf{Z}^{(2)} \| \mathbf{U}] \right) \odot \mathbf{Z}^{(2)} 
+ \Big(1 - \sigma\!\left(\mathbf{\Tilde{W}} \,[\mathbf{Z}^{(2)} \| \mathbf{U}] \right)\Big) \odot \mathbf{U}
\end{equation}
where $\sigma(\cdot)$ denotes the element-wise sigmoid function, which dynamically generates a soft gate between the local and global information for each node. 
$\mathbf{\Tilde{XW}} \in \mathbb{R}^{2d\times d}$ is a learnable linear transformation that projects the concatenated features back to $\mathbb{R}^{b\times T\times d}$. The operator $\left |  \right | $ is the concatenation of features along the channel dimension. In addition, the fused feature $\mathbf{\Tilde{X}}$ is computed as a weighted combination of $\mathbf{H}^{(2)}$ and $\mathbf{U}$, enabling adaptive emphasis on local and global semantics.

Finally, we apply a fully connected layer to obtain the updated node features. Noted that we place the CATG module into audio and visual branch separately and share the parameters with each other. 
\begin{algorithm}[t]
\small
\caption{Category-Aware Temporal Graph Construction}
\label{alg:cats-construct}
\begin{algorithmic}[1]
\Require $\mathbf{X} \in \mathbb{R}^{B\times T\times d}$, 
         $\mathbf{P} \in \mathbb{R}^{B\times T\times C}$, 
         candidate hop set $\mathcal{K}=\{1,\ldots,K\}$, 
         temperature $\tau$, 
         hyper-parameter $k$,
parameters (learnable, trainable via back-propagation): 
         $\mathbf{W}_K \in \mathbb{R}^{C\times K}$, 
         $\mathbf{w}_\lambda \in \mathbb{R}^{C}$
\Ensure $\mathcal{G}=(\mathcal{N},\mathcal{E},\mathbf{W})$
\State $\mathcal{N} \gets \{1,\ldots,T\}$;\; $\mathcal{E} \gets \varnothing$;\; $\mathbf{W} \gets \mathbf{0}\in \mathbb{R}^{B\times T\times T}$
\State $\mathbf{H} \gets \mathbf{P} \mathbf{W}_K \in \mathbb{R}^{B\times T\times K}$ \Comment{hop preference}
%\For{$b=1$ to $B$}
\For{$t=1$ to $T$}
  \State $\mathcal{E}\gets \mathcal{E}\cup \{(t,t)\}$;\; ${\mathbf{w}_{t,t}\gets \mathbf{1}\in \mathbb{R}^{B}}$
  \State $\mathbf{s}_{t,i} \gets \dfrac{\exp((\mathbf{h}_{t,i}+\mathbf{n}_{t,i})/\tau)}{\sum_{j=1}^K \exp((\mathbf{h}_{t,j}+\mathbf{n}_{t,j})/\tau)}$, $\forall i \in \mathcal{K}$
  \State $\mathcal{L}_t \gets \{l_1,\ldots,l_k\} \gets \mathrm{Top}\text{-}k(\{s_{t,i}\}_{i=1}^K)$
  \State $\mathbf{q}_t \gets \mathbf{P}_{t} \mathbf{w}_\lambda$
  \For{each $l_i \in \mathcal{L}_t$ and $t+l_i \le T$}
    \State $\mathcal{E} \gets \mathcal{E} \cup \{(t,\,t{+}l_i)\}$
    \State $\mathbf{w}_{t,\,i} \gets \mathbf{s}_{t,i}\odot\exp(-\mathbf{q}_t l_i)$
    %w_ = CNN {dsfdsj}
    %w = 2p
  \EndFor
\EndFor
\State Assemble the adjacency matrix $\mathbf{W}$ where each entry 
$\mathbf{W}_{b,\,t,\,t{+}l_i} = \mathbf{w}_{t,i}$,
and $\mathbf{w}_{t,i}$ is computed using the learnable parameters 
$\mathbf{W}_K$ and $\mathbf{w}_\lambda$ as:
\[
\mathbf{w}_{t,i} = \mathrm{Softmax}\!\left(\mathbf{P}_t \mathbf{W}_K\right)
\cdot \exp(-\mathbf{w}_\lambda \, l_i),
\quad
\mathbf{W}\in\mathbb{R}^{B\times T\times T}.
\]
\State \textbf{Optimization:} 
Gradients of $\mathbf{W}_K$ and $\mathbf{w}_\lambda$ are obtained 
through back-propagation from the overall task loss $\mathcal{L}_{\text{task}}$ 
of our network (since the graph construction is fully differentiable 
in Eqs.~(5)--(8)).
Both parameters are updated via gradient descent:
\[
\mathbf{W}_K \leftarrow \mathbf{W}_K - \eta \frac{\partial \mathcal{L}_{\text{task}}}{\partial \mathbf{W}_K}, 
\quad
\mathbf{w}_\lambda \leftarrow \mathbf{w}_\lambda - \eta \frac{\partial \mathcal{L}_{\text{task}}}{\partial \mathbf{w}_\lambda}.
\]
\State \Return $\mathcal{G}=(\mathcal{N},\mathcal{E},\mathbf{W})$
\end{algorithmic}
\end{algorithm}
\vspace{-4mm}
\subsection{Gated Fusion}
To fully exploit the complementary advantages of the feature aggregation and graph reasoning branches, we introduce a gated fusion module that dynamically balances the contributions of HAN-aggregated and CATG-refined features. 

We consider that HAN aggregates audio or visual features into temporally-aware semantic representations by emphasizing salient segments via attention mechanisms. This results in $\mathbf{M}_{HAN}\in\mathbb{R}^{b\times T\times d}$, which captures modality-specific semantic importance and local temporal context. Building upon this, CATG module refine the feature representation into $\mathbf{M}_{CATG}\in\mathbb{R}^{b\times T\times d}$ to model long-range dependencies, time-skipped correlations, and event-level structure. Although $\mathbf{M}_{CATG}$ is derived from $\mathbf{M}_{HAN}$, the graph propagation process transforms the feature space by emphasizing structural consistency and relational context. Hence, the two outputs are semantically related but express different views of the input, which are variably useful across segments. 

Therefore, we design a gated fusion mechanism to adaptively combine the HAN and CATG outputs at the segment level. Specifically, we first concatenate the features $\mathbf{M}_{HAN}$ and $\mathbf{M}_{CATG}$ along the feature dimension, and use a gating function to calculate the segment-wise fusion weights $\mathbf{G}$:
\begin{equation}
\label{gated_fusion_2}
\mathbf{O} = \sigma \big( \mathbf{\Bar{W}} [\mathbf{M}_{HAN}; \mathbf{M}_{CATG}] + \mathbf{\Bar{b}} \big), \quad 
\mathbf{O} \in \mathbb{R}^{b\times T\times d}
\end{equation}
where $\mathbf{\Bar{W}}\in\mathbb{R}^{2d\times d}$ and $\mathbf{\Bar{b}}\in\mathbb{R}^{d}$ are the learnable weight matrix and bias of a linear transformation. The final
fused representation $\mathbf{F}_{Fuse}$ is computed as:
\begin{equation}
\label{gated fusion_2}
\mathbf{M}_{Fuse} = \mathbf{O}\odot \mathbf{M}_{HAN}+(\mathbf{1}-\mathbf{O})\odot \mathbf{M}_{CATG}
\end{equation}
where $\odot$ denotes element-wise multiplication, and $\mathbf{1}$ is a tensor of the same dimension as $\mathbf{O}$, with all elements equal 1. This gating mechanism enables the model to dynamically favor either the semantic-aware or structure-aware representation at each temporal location, depending on their relative informativeness. It serves as a critical bridge between semantic reasoning and temporal-structural inference in our framework.
\vspace{-6mm}
\section{Experiments}
\subsection{Experimental Setup}
\subsubsection{Dataset}
We first evaluate our framework on the LLP dataset~\cite{tian2020unified}, which contains 11,849 videos annotated with 25 daily-life event classes. It provides weak labels for 10,000 training videos, fully annotated labels for 649 videos for validation, and 1,200 videos for testing. Each video is divided into 10 segments, each of one second.
We further perform testing on the UnAV-100 dataset~\cite{geng2023dense}, a large-scale benchmark for dense audio-visual event localization, containing 10,790 videos and over 30,000 event instances across 100 event categories. Videos vary from 0.2 to 60 seconds, often with multiple concurrent events. Following CoLeaF~\cite{sardari2024coleaf}, we adopt a weakly supervised setting by using only video-level labels for training.

\subsubsection{Implementation Details}
For the LLP dataset, we follow VALOR~\cite{lai2023modality} by using the pre-trained CLAP~\cite{wu2023large} model to extract 768-dimensional audio features from the audio signal. For the visual modality, we employ pre-trained CLIP~\cite{radford2021learning} and 3D ResNet to extract 768-dimensional and 512-dimensional features. 
In addition, we train our framework using the batch size of 128 for 10 epochs. 
For the parameters of the CATG module, we use the candidate skip lengths $k\in\{1,2,...,9\}$, where top-3 skip lengths are dynamically selected via Gumbel-Softmax with a temperature of 1.0.

For the UnAV-100 dataset, we follow~\cite{geng2023dense} to extract 2048 dimensional visual features using a two-stream I3D model (RGB + RAFT), and obtain 128 dimensional audio features from pre-trained VGGish aligned with 0.32s video segments~\cite{teed2020raft,hershey2017cnn}. Since UnAV-100 lacks CLIP/CLAP features, we focus solely on evaluating the CATG module. As a self-contained design that does not rely on pseudo labels, this setup enables a fair assessment of CATG's temporal modeling capacity under large-scale weak supervision. Evaluation of the BiT module is left to the LLP dataset, where high-quality pseudo labels are available.
\subsubsection{Evaluation Metrics}
For the LLP datset, we adopt the F1-score as the evaluation metric for three types of events: audio (\textbf{A}), visual (\textbf{V}), and audio-visual (\textbf{AV}). We use a fixed threshold of mIoU = 0.5 to determine positive predictions. Evaluation is conducted at both the segment level and the event level. At the segment level, predictions are compared with ground truth labels on a per-segment basis. At the event level, multiple contiguous segments predicted as the same event are grouped together and evaluated as a single event instance. In addition, the metric $\textbf{Type{@}AV}$ calculates the average F1-score over the three event types, while $\textbf{Event{@}AV}$ reflects the overall F1-score by jointly considering all audio and visual events present in each video. We also use \textbf{AV} F-score to evaluate the effect on the UnAV-100 dataset.
\subsubsection{Baseline Methods}
To show the effectiveness of our model, we compare our TEn-CATG with several other baselines on two different datasets, including advanced structural methods such as MA~\cite{wu2021exploring}, CMPAE~\cite{gao2023collecting}, and CoLeaF~\cite{sardari2024coleaf}, and pseudo-label-based methods such as VALOR~\cite{lai2023modality}, LSLD~\cite{fan2024revisit}, and NREP~\cite{jiang2024resisting}. 
\begin{table*}[t]
  \caption{The performance of TEn-CATG and comparative methods in AVVP, with the best results highlighted in \ \cellcolor{c2!80}{\textbf{bold}} and the second results highlighted in \cellcolor{c1!40}{text}.}
  \label{tab:main-results}
  \centering
  \renewcommand{\arraystretch}{1.1}
  % \begin{tabular}{l|c|ccccc|ccccc}
  \begin{tabular}{l|c|ccccc|ccccc|c}
    \hline
    % \multirow{2}{*}{Model} & \multirow{2}{*}{Venue} & \multicolumn{5}{c|}{Segment-level (\%)} & \multicolumn{5}{c}{Event-level (\%)}\\
    \multirow{2}{*}{Model} & \multirow{2}{*}{Venue} & \multicolumn{5}{c|}{Segment-level (\%)} & \multicolumn{5}{c|}{Event-level (\%)} & \multirow{2}{*}{Avg~(\%)} \\
    \cline{3-7} \cline{8-12}
    &  & A & V & AV & Type@AV & Event@AV & A & V & AV & Type@AV & Event@AV \\
    \hline
    HAN~\cite{tian2020unified} & ECCV'20 & 60.1 & 52.9 & 48.9 & 54.0 & 55.4 & 51.3 & 48.9 & 43.0 & 47.7 & 48.0 & 51.0\\
    % \cmidrule{2-19}
    MGN~\cite{mo2022multi} & NeurIPS'22 & 60.8 & 55.4 & 50.0 & 55.1 & 57.6 & 52.7 & 51.8 & 44.4 & 49.9 & 50.0 & 52.8\\
    % \cmidrule{2-19}
    MA~\cite{wu2021exploring} & CVPR'21 & 60.3 & 60.0 & 55.1 & 58.9 & 57.9 & 53.6 & 56.4 & 49.0 & 53.0 & 50.6 & 55.5 \\
    % \cmidrule{2-19}
    CMPAE~\cite{gao2023collecting} & CVPR'23 & 64.2 & 66.2 & 59.2 & 63.3 & 62.8 & 56.6 & 63.7 & 51.8 & 57.4 & 55.7 & 60.1\\
    % \cmidrule{2-19}
    CoLeaF~\cite{sardari2024coleaf} & ECCV'24 & 64.2 & 67.1 & 59.8 & 63.8 & 61.9 & 57.1 & 64.8 & 52.8 & 58.2 & 55.5 & 60.5\\
    % \cmidrule{2-19}
    LEAP~\cite{zhou2024label} & ECCV'24 & 64.8 & 67.7 & 61.8 & 64.8 & 63.6 & 59.2 & 64.9 & \cellcolor{c1!40}{56.5} & 60.2 & 57.4 & 62.1\\
    % \cmidrule{2-19}
    VALOR++~\cite{lai2023modality} & NeurIPS'23 & 68.1 & 68.4 & 61.9 & 66.2 & 66.8 & 61.2 & 64.7 & 55.5 & 60.4 & 59.0 & 63.2\\
    % \cmidrule{2-19}
    LSLD+~\cite{fan2024revisit} & NuerIPS'23 & 68.7 & \cellcolor{c1!40}{71.3} & \cellcolor{c1!40}{63.4} & 67.8 & 68.2 & \cellcolor{c1!40}{61.5} & \cellcolor{c1!40}{67.4} & 55.9 & 61.6 & 60.6 & 64.6\\
    % \cmidrule{2-19}
    NREP~\cite{jiang2024resisting} & TNNLS'24 & \cellcolor{c1!40}{70.2} & 70.9 & \cellcolor{c2!80}\textbf{64.4} & \cellcolor{c1!40}{68.5} & \cellcolor{c1!40}{68.8} & \cellcolor{c2!80}\textbf{62.8} & 67.3 & \cellcolor{c2!80}\textbf{57.6} & \cellcolor{c2!80}\textbf{62.6} & \cellcolor{c1!40}{61.1} & \cellcolor{c1!40}{65.4}\\
    \hline
    TEn-CATG~(Ours) & - & \cellcolor{c2!80}\textbf{73.7}  & \cellcolor{c2!80}\textbf{74.1}  & 63.2 & \cellcolor{c2!80}\textbf{70.3} & \cellcolor{c2!80}\textbf{73.9} & {61.1} & \cellcolor{c2!80}\textbf{70.3} & 54.3 & \cellcolor{c1!40}\textbf{61.9} & \cellcolor{c2!80}{\textbf{61.9}} & \cellcolor{c2!80}\textbf{66.5} \\
    &  & \textcolor{blue}{(\textbf{+3.5})} & \textcolor{blue}{(\textbf{+2.8})} &  & \textcolor{blue}{(\textbf{+1.8})} & \textcolor{blue}{(\textbf{+5.1})} &  & \textcolor{blue}{(\textbf{+2.9})} &  & & \textcolor{blue}{(\textbf{+0.8})}&\textcolor{blue}{(\textbf{+1.1})} \\
  \hline
\end{tabular}
% }
\end{table*}
\vspace{-3mm}
\subsection{Overall Performance Analysis}
\subsubsection{Comparative Performance Overview}
As shown in Table~\ref{tab:main-results}, TEn-CATG achieves significant improvements over prior methods across both segment-level and event-level metrics. Particularly at the segment level, our model demonstrates its superior ability to model fine-grained temporal and semantic structures under weak supervision.

The notable gain in segment-level Event@AV reflects the synergy between our BiT and CATG modules. This shows that the model can propagate consistent semantic signals across time and suppress spurious activations, which is particularly effective when events are temporally fragmented or weak in one modality. Together, these two modules enable the model to identify and localize audio-visual events more accurately at the segment level.

In contrast, our method is slightly lower than some baselines in the event-level AV prediction (54.3\%). We attribute this to the conservative fusion and pooling strategy, which emphasizes precision by aggregating only confident segment predictions. While this helps avoid false positives, it may reduce recall for low-saliency or asynchronous events. Additionally, noisy or incomplete pseudo labels may further degrade fusion quality at the video level.

Nevertheless, our model achieves the highest segment-level Event@AV and competitive performance in event-level Type@AV (61.9\%), confirming the effectiveness of our framework in learning semantically aligned and structurally coherent patterns. These results validate our core motivation: modeling semantic relationships across segments and aligning modality features through unified textual guidance yields robust performance beyond standard modality fusion.
\begin{figure}[t] 
\centering
\includegraphics[width=0.45\textwidth]{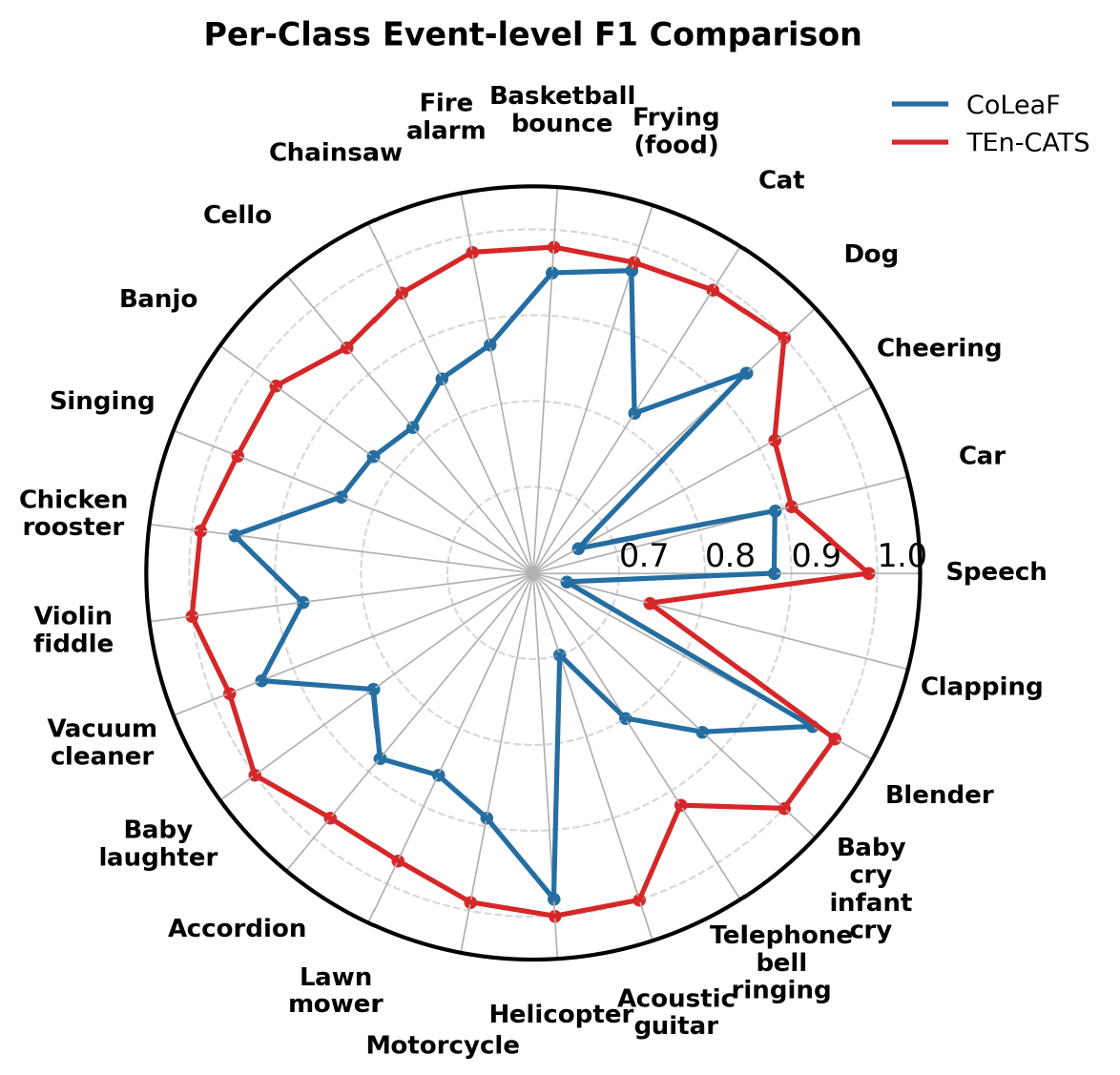} % 调整图片宽度
    \caption{Per-class event-level F1 score comparison between CoLeaF (blue) and our model (red).} 
     % Our method consistently outperforms the baseline across most event categories, particularly on previously low-performing classes such as Cheering and Clapping.
    \label{fig:event f1}
\end{figure} 
\subsubsection{Event-Wise Discriminative Power Performance}
To further understand the performance gains of our model, we conduct a per-class event-level F1 score comparison against the baseline CoLeaF~\cite{sardari2024coleaf}. As illustrated in Fig.~\ref{fig:event f1}, our model consistently achieves higher per-class F1 scores across nearly all event categories. Notably, substantial improvements are observed in previously underperforming classes in CoLeaF, such as Cheering (from 0.66 to 0.92), Acoustic\_guitar (from 0.70 to 0.97), and Telephone\_bell\_ringing (from 0.80 to 0.92), which are typically characterized by transient and overlapping audio-visual events. The improved recognition of these fine-grained events demonstrates the model’s enhanced ability to disentangle semantically coupled cues.

Despite the fact that most F1 scores achieved by our model exceed 0.90, the metric of event-level Event@AV in Table~\ref{tab:main-results} is 61.9\%. This gap is because Fig.~\ref{fig:event f1} reflects the macro F1, treating each category equally regardless of its frequency. While the Event@AV is micro F1, which is dominated by high-frequency events, such as Speech and Singing. Even slight underperformance in these dominant classes can significantly affect the global metric.

Nevertheless, the alignment between class-wise consistency (Fig.~\ref{fig:event f1}) and strong overall results (Table~\ref{tab:main-results}) confirm the robustness of our model across both frequent and rare event categories. These fine-grained improvements suggest that our temporal and semantic modeling effectively enhances per-category discrimination while maintaining competitive global performance.

\subsubsection{Cross-Modal Semantic Separability Visualization}
\begin{figure}[t] 
\centering
\includegraphics[width=0.5\textwidth]{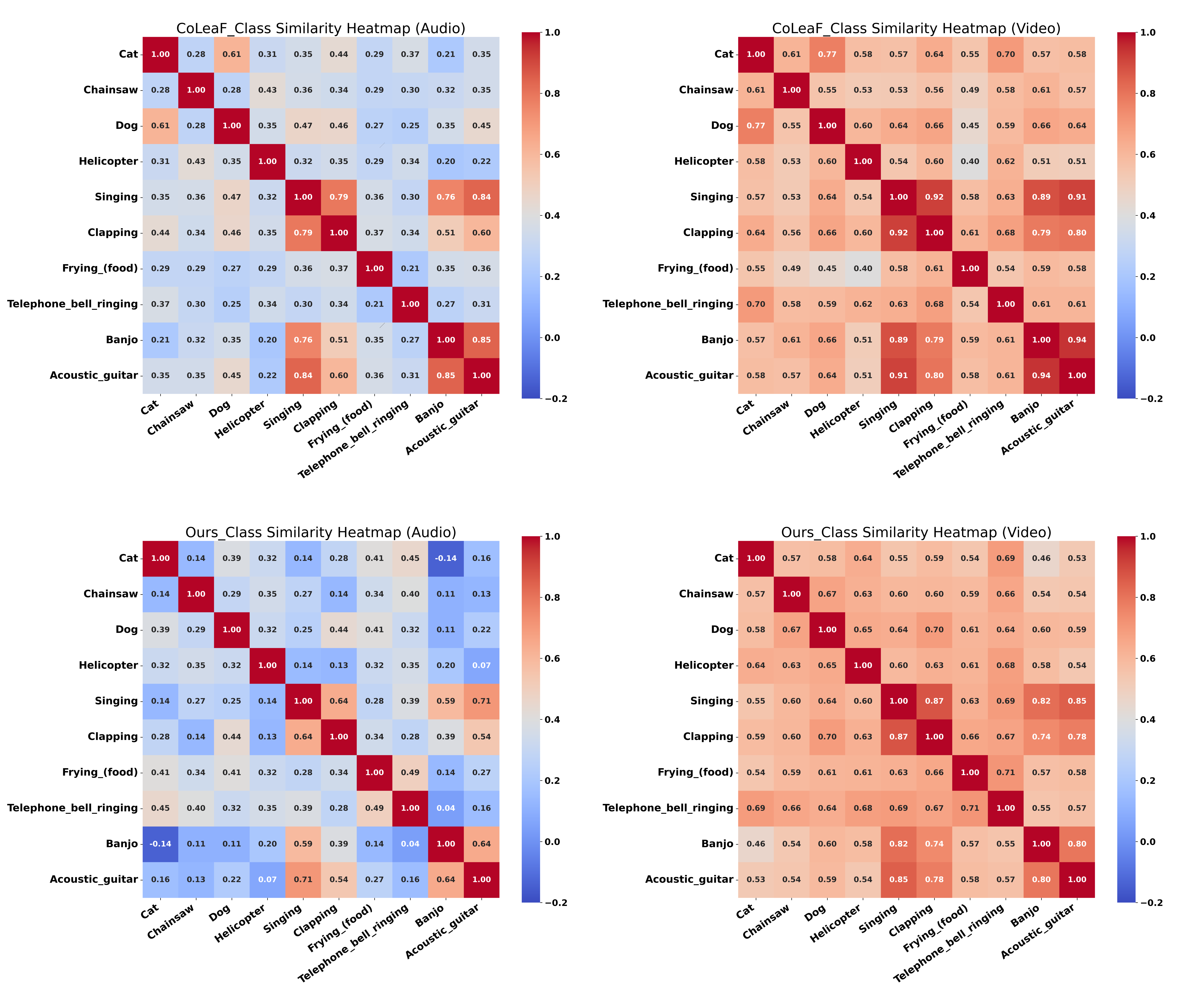} % 调整图片宽度
    \caption{Per-class cosine similarity heatmaps of audio~(left) and visual~(right) features for CoLeaF~(top) and our model~(bottom).}
    % Our model demonstrates improved intra-class compactness and inter-class separability, especially in the audio modality, highlighting better semantic structure learning.
    \label{fig:cos}
\end{figure}

To analyze the semantic structure of learned features, we visualize the per-class cosine similarity matrices for the audio and visual modalities as in Fig.~\ref{fig:cos}. A higher off-diagonal similarity indicates a higher inter-class confusion and a poorer modality-specific separability.

In the audio heatmap, CoLeaF shows high similarity between semantically distinct classes (e.g., Singing vs. Clapping: 0.79; Acoustic\_guitar vs. Banjo: 0.85), indicating poor discriminability. In contrast, our model yields lower similarity for these pairs (e.g., both 0.64), suggesting clearer semantic boundaries.
For visual features, both models maintain strong intra-class similarity, but our model reduces confusion between highly similar classes. For instance, Banjo versus Acoustic\_guitar drops from 0.94 to 0.80, and Clapping versus Singing from 0.92 to 0.87. These reductions align with per-class F1 gains and confirm our model’s ability to learn better separated, semantically structured features.

\begin{figure*}[h]
    \centering    \includegraphics[width=0.93\textwidth]{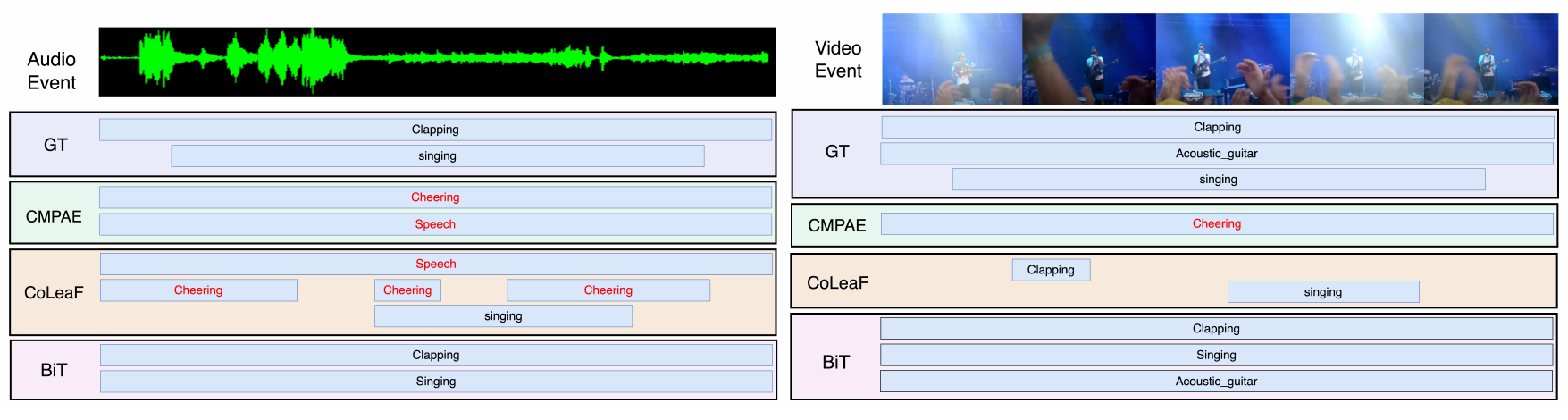} % 调整图片宽度
    \caption{Qualitative comparison of audio and video event predictions on a sample video. The BiT-only model (Ours) achieves more accurate and complete predictions compared to CMPAE and CoLeaF, closely matching the ground truth (GT) for both modalities.}
    \label{fig:bit-ablation}
\end{figure*}
\vspace{-4mm}
\subsection{Ablation Study}
\subsubsection{Ablation Study on Module Contribution}
To verify the contribution of the proposed BiT and CATG modules, we conduct ablation studies by disabling these two components in TEn-CATG. To ensure a fair comparison with CoLeaF, which was originally built upon audio and visual features extracted by VGGish and ResNet, we re-train CoLeaF using the same input features as TEn-CATG, denoted as CoLeaF\textsuperscript{†}. To validate the effectiveness and individual contribution of each proposed module, we first conduct ablation studies on the LLP dataset. Specifically, we compare the performance of TEn-CATG when removing or isolating the BiT and CATG modules, as shown in Table~\ref{tab:abs-results}.

Overall, we observe that the best results for A and Event@AV at both segment and event levels are achieved when only the BiT module is applied, while the best results for V, AV, and Type@AV at both levels are obtained when combining both BiT and CATG (full model). This reveals a clear trade-off of our model between enhancing unimodal performance, especially audio-driven event recognition, and maximizing overall multi-modal and event consistency. For the best performance in A and Event@AV when only adding the BiT module, this is probably because the BiT-only configuration focuses on strengthening the semantic alignment between modality-specific features and event concepts without introducing additional temporal constraints. This simplifies the learning objective, allowing the model to better capture highly distinguishable audio characteristics, such as speech patterns, musical rhythms, environmental sounds (e.g., alarm, dog barking), or other acoustically unique event cues.
These audio events are often globally consistent across segments, making them easier to capture through semantic enhancement alone, without requiring complex temporal reasoning.
Moreover, this semantic reinforcement directly benefits holistic event detection (Event@AV) by helping the model capture high-level event semantics across segments, even if temporal precision is not fully optimized.

Focusing on the performance of our framework, the full model introduces category-aware temporal modeling, which is particularly effective for visual streams, where events often happen with clear temporal structures (e.g., object actions, scene changes).
The combination of semantic and temporal cues also benefits AV fusion and type-level performance (Type@AV), as it helps the model to align and balance both modalities over time. However, this more complex modeling may slightly dilute unimodal audio strength, as the model focuses more on multi-modal consistency rather than maximizing single-modality performance.

\begin{table}[htpb]
  \caption{Ablation study for TEn-CATG. \textit{o.} means only have certain module, \textit{A} and \textit{V} means audio branch and visual branch.}
  \label{tab:abs-results}
  \centering
  \setlength{\tabcolsep}{1.2mm}{
  \renewcommand{\arraystretch}{1.1}
  \begin{tabular}{clccccc}
    \hline
    % & & \multicolumn{5}{c}{Segment-level (\%)}\\
    % \midrule
    & Method & A & V & AV & Type@AV & Event@AV\\
    \hline
   \multirow{6}{*}{Segment-level} & CoLeaF$^\dagger$ & 64.2  & 67.4  & 59.9 & 63.8 & 63.3\\
    & CATG~\textit{o.~A} & 65.5  & 68.1  & 59.6 & 64.4 & 64.5\\
    & CATG~\textit{o.~V} & 66.0  & 68.9  & 60.4 & 65.1 & 65.2\\
   & CATG full & 65.3 & 68.7 & 60.8 & 64.9 & 64.6 \\
   & \textit{o.}~BiT & \textbf{74.2 }& 73.4 & 62.5 & 70.1 & \textbf{74.1} \\
    \cline{2-7}
    & TEn-CATG & 73.7 & \textbf{74.1} & \textbf{63.2} & \textbf{70.3} & 73.9\\
   \hline
    % & & \multicolumn{5}{c}{Event-level (\%)}\\
    % \midrule
   & Method & A & V & AV & Type@AV & Event@AV\\
    \hline
   \multirow{6}{*}{Event-level} & CoLeaF$^\dagger$ & 53.2 & 64.1 & 52.4 & 56.6 & 52.7 \\
   & CATG~\textit{o.~A} & 54.5  & 64.8  & 51.5 & 56.9 & 54.0\\
   & CATG~\textit{o.~V} & 55.1  & 65.4  & 52.2 & 57.6 & 54.7\\
    & CATG full & 54.7 & 64.8 & 52.8 & 57.5 & 54.3\\
    & \textit{o.}~BiT & \textbf{61.6} & 69.5 & 53.5 & 61.5 & \textbf{62.1}\\
   \cline{2-7}
   & TEn-CATG &61.1 & \textbf{70.3} & \textbf{54.3} & \textbf{61.9} & 61.9\\
  \hline
\end{tabular}
}
\end{table}

\subsubsection{Qualitative Case Study on the BiT Module}
To assess the impact of the BiT module, we visualize a representative example from the LLP dataset in Fig.~\ref{fig:bit-ablation}. This case illustrates the predicted temporal segments of the audio and visual events by different models, including CMPAE~\cite{gao2023collecting}, CoLeaF~\cite{sardari2024coleaf}, and our model with only the BiT module enabled.

On the audio event timeline, the ground truth consists of two co-occurring events, Clapping and Singing. CMPAE and CoLeaF both misclassify segments as Cheering or Speech, failing to capture the correct event boundaries or producing semantically unrelated predictions. In contrast, our BiT-only model accurately detects both Clapping and Singing. Although the model also predicts Acoustic\_guitar~(a false positive not present in the ground truth), this may stem from the model's semantic sensitivity to background instrument cues, which are audible in the clip. On the video event timeline, a similar trend is observed. While CMPAE and CoLeaF struggle with incorrect predictions (e.g., false positive Cheering in CMPAE, fragmented Singing in CoLeaF), the BiT-only model produces temporally aligned and semantically correct predictions for Clapping, Singing, and Acoustic\_guitar. This suggests that the BiT module provides strong semantic priors, enabling the model to better associate event concepts with modality-specific patterns even in the absence of explicit temporal modeling.

\subsubsection{Analysis of Temporal Scale in the CATG Module}
\begin{figure}[t] 
\centering
\includegraphics[width=0.5\textwidth]{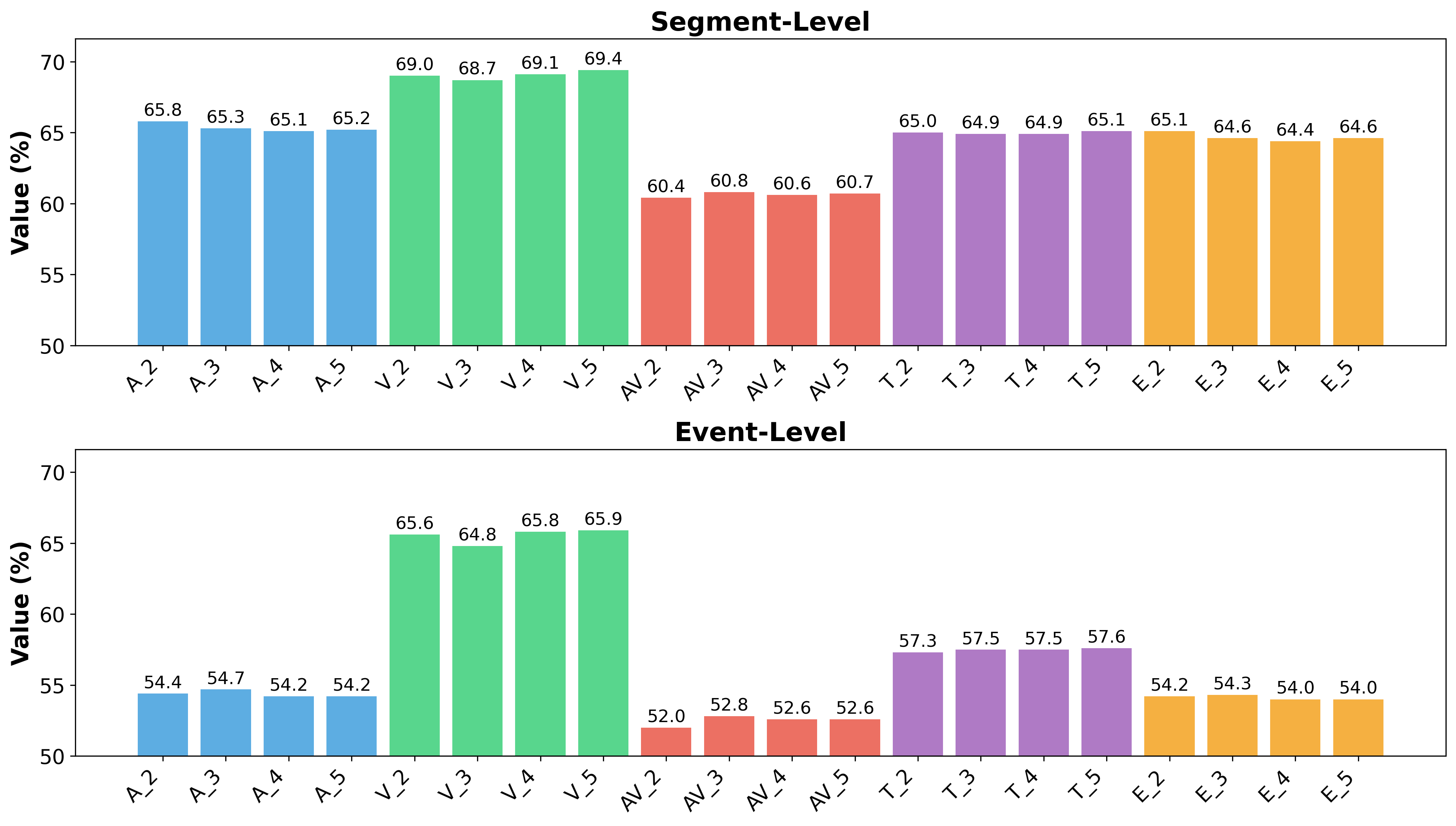} % 调整图片宽度
    \caption{Segment- and event-level performance of the CATG module under different hop sizes $k$. Each color represents one evaluation metric~(A, V, AV, Type@AV, Event@AV).} 
    \label{fig:cats-k}
\end{figure}

To evaluate the sensitivity of our CATG module to the choice of the temporal hop size $k$, we conduct controlled experiments by varying $k\in{2,3,4,5}$. Fig.~\ref{fig:cats-k} presents the performance across five key metrics (A, V, AV, Type@AV, Event@AV) at both the segment-level and event-level, color-coded for each metric group.

At the segment-level, we observe that increasing $k$ from 2 to 3 leads to a consistent improvement across all metrics, with the visual branch (V) reaching its peak at 69.1\% for $k=3$. This suggests that moderate temporal hops are beneficial for modeling visual temporal patterns, which often depend on structured changes such as object movement or scene transitions. The audio branch (A), by contrast, shows less variation across $k$, indicating its robustness to temporal granularity due to the inherently high temporal resolution of audio features.

The AV, Type@AV, and Event@AV metrics also achieve their best or near-best performance at $k=3$ or $k=4$, with a noticeable drop at $k=2$. This suggests that too short temporal hops limit the model's ability to capture meaningful event transitions or category evolution patterns. At the same time, large hop sizes, such as $k=5$ lead to slight performance degradation in some metrics, potentially due to over-smoothing or temporal misalignment.

At the event-level, a similar trend emerges. While audio branch performance (A) remains relatively stable, the visual branch (V) again peaks at $k=3$, achieving 65.9\%, and the multi-modal metrics (AV, Type@AV, Event@AV) favor mid-range values $k=3$ and $k=4$. Notably, AV performance rises from 52.0\% at $k=2$ to 57.6\% at $k=5$, but the improvement plateaus after $k=4$, suggesting diminishing returns with longer temporal spans.

These results highlight the importance of carefully selecting the temporal scale in temporal graph construction. Specifically, a moderate hop size (e.g., $k=3$) offers the best trade-off between capturing meaningful temporal dependencies and maintaining alignment across segments. 

\begin{table}[ht]
    \caption{Comparison of CATG performance on the weakly-labeled UnAV-100 dataset.}
    \label{UnAV100}
\centering
\begin{tabular}{lcc}
\toprule
\textbf{Method} & \textbf{AV (Seg)} & \textbf{AV (Evn)} \\
\midrule
HAN~\cite{tian2020unified}     & 35.0 & 41.4 \\
MA~\cite{wu2021exploring}       & 37.9 & 44.8 \\
JoMoLD~\cite{cheng2022joint} & 36.4 & 41.2 \\
CMPAE~\cite{gao2023collecting} & 39.7 & 43.8 \\
CoLeaF\cite{sardari2024coleaf} & 41.5 & \textbf{47.8} \\
\midrule
\textbf{CATG} & \textbf{41.9} \textcolor{blue}{\textbf{(+0.4)}} & \underline{47.5}\\
\bottomrule
\end{tabular}
\end{table}

\subsubsection{Effectiveness of CATG on UnAV-100}
To evaluate the generalizability and robustness of our proposed CATG module, we conduct additional experiments on the UnAV-100 dataset using the weak labels provided by CoLeaF. As shown in Table~\ref{UnAV100}, the incorporation of CATG yields further improvement over the previous state-of-the-art methods.

Compared to CoLeaF, our method achieves 41.9\% segment-level mAP and 47.5\% event-level mAP in the AV branch, surpassing it by +0.4\% and maintaining competitive performance on the event level. This gain, albeit marginal, is consistent and meaningful under weak supervision, demonstrating that CATG can effectively model category-aware temporal structures even when ground-truth timestamps are unavailable.

In particular, the improvement on the segment-level metric highlights CATG’s ability to enhance temporal localization and reduce false positives in ambiguous segments. These results further validate that our framework generalizes well across different weakly labeled datasets and benefits from the fine-grained temporal reasoning introduced by the CATG module.
\vspace{-8mm}
\section{Conclusion}
We have presented a novel framework for AVVP task by designing the BiT and CATG modules. 
Extensive experiments on LLP and UnAV-100 datasets demonstrate the effectiveness of our approach. BiT enhances audio-visual understanding, especially for ambiguous or rare events, while CATG ensures consistent semantic reasoning over time. Ablation studies further validate the individual contributions and the strong complementarity of both modules.
Our model achieves new state-of-the-art results and shows robust generalization under real-world conditions. Future work includes handling overlapping events, enhancing robustness of the models to label noise, and extending to tasks such as audio-visual instance segmentation and cross-modal retrieval.
\vspace{-4mm}
\section*{Acknowledgments}
This work was partially supported by a research scholarship from the China Scholarship Council (CSC) and a studentship from the University of Surrey. For the purpose of open access, the authors have applied a Creative Commons Attribution (CC BY) license to any Author Accepted Manuscript version arising. 
\vspace{-1.2em}
\bibliographystyle{IEEEtran}
\bibliography{sample-base}

\end{document}